\documentclass[10pt,twocolumn,letterpaper]{article}

\usepackage{cvpr}              

\usepackage{graphicx}
\usepackage{amsmath}
\usepackage{amssymb}
\usepackage{booktabs}
\usepackage{comment}
\usepackage[dvipsnames]{xcolor}
\usepackage{color, colortbl}
\usepackage[accsupp]{axessibility}

\usepackage[pagebackref,breaklinks,colorlinks]{hyperref}

\usepackage[capitalize]{cleveref}
\crefname{section}{Sec.}{Secs.}
\Crefname{section}{Section}{Sections}
\Crefname{table}{Table}{Tables}
\crefname{table}{Tab.}{Tabs.}

\def\cvprPaperID{56} 
\def\confName{CVPR}
\def\confYear{2022}

\begin{document}

\title{Exposure Correction Model to Enhance Image Quality}

\author{Fevziye Irem Eyiokur\textsuperscript{1,*} \qquad Dogucan Yaman\textsuperscript{1,*} \qquad Hazım Kemal Ekenel\textsuperscript{2} \qquad Alexander Waibel\textsuperscript{1,3}\\
\textsuperscript{1}Karlsruhe Institute of Technology, \textsuperscript{2}Istanbul Technical University, \textsuperscript{3}Carnegie Mellon University\\
{\tt\small \{irem.eyiokur, dogucan.yaman, alexander.waibel\}@kit.edu, ekenel@itu.edu.tr}
}
\maketitle
\renewcommand{\thefootnote}{\Alph{footnote}}
\begin{NoHyper}\footnotetext{\textsuperscript{*}Equal contribution.}\end{NoHyper}

\begin{abstract}
Exposure errors in an image cause a degradation in the contrast and low visibility in the content. In this paper, we address this problem and propose an end-to-end exposure correction model in order to handle both under- and overexposure errors with a single model. Our model contains an image encoder, consecutive residual blocks, and image decoder to synthesize the corrected image. We utilize perceptual loss, feature matching loss, and multi-scale discriminator to increase the quality of the generated image as well as to make the training more stable. The experimental results indicate the effectiveness of proposed model. We achieve the state-of-the-art result on a large-scale exposure dataset. Besides, we investigate the effect of exposure setting of the image on the portrait matting task. We find that under- and overexposed images cause severe degradation in the performance of the portrait matting models. We show that after applying exposure correction with the proposed model, the portrait matting quality increases significantly. 
\textcolor{magenta}{\href{https://github.com/yamand16/ExposureCorrection}{\textit{https://github.com/yamand16/ExposureCorrection}}}.
\end{abstract}



\section{Introduction}

The quality of the images depends on several factors and dramatically affects the performance of the computer vision methods. 
The exposure attribute is one of these factors and depends on shutter speed, f-number, and camera ISO. The exposure setting is expressed by exposure values (EVs) and each EV yields a different level of brightness in the image. In the case that zero EV value is the proper setting for an arbitrary image, negative EV makes it underexposed, while positive EV causes overexposed version. Besides, underexposed images have a darker appearance and overexposed images have a brighter view. Moreover, both situations cause low visibility.  
Therefore, exposure correction is a key step to overcome exposure errors to provide a better image.

The main goal of the exposure correction is to adjust the exposure setting of an image to generate the same image with better content visibility, appropriate brightness level, and a more clear appearance. While doing this, one should be careful not to deform the content as well as the color distribution and should avoid inducing noise. 
In the literature, the exposure correction task is addressed by various methods. First of all, generic image quality enhancement~\cite{gharbi2017deep,chen2018deep,ouyang2021neural} and relighting~\cite{el2021ntire} are employed to adjust lighting in order to improve the quality of the image. Furthermore, low-light image enhancement~\cite{huang2022towards,lore2017llnet,zhang2021beyond,lime2,DeepUPE,zhang2018high,KinD,xu2020learning,yang2020fidelity,zhu2020eemefn,yu2018deepexposure,zhang2019dual} is proposed to directly deal with the exposure problem. However, all these methods treat the exposure correction for either under- or overexposed images. Recently, novel methods~\cite{afifi2021learning,yang2022learning,ouyang2021neural,liang2022fusion} are successfully handled the under- and overexposure problem at the same time. 



In this paper, we address the exposure problem and propose a generative adversarial network-based exposure correction model to adjust the exposure setting of an image in order to enhance it. 
For this, our model receives an input image and encodes it with the image encoder to provide latent representation. Then, this representation is processed by a residual block to edit the features and passes through the image decoder to synthesize the final image. 
We comprehensively analyze the performance on the large-scale exposure correction dataset~\cite{afifi2021learning} and achieve the state-of-the-art results. Additionally, we investigate the effect of exposure error on a real-world application, namely, portrait matting. 
To perform this, we choose four real-world portrait matting datasets and manipulate them by Adobe Photoshop Lightroom to obtain under- and overexposed images. After, we run the SOTA portrait matting models on these images. Later, we apply exposure correction with our model and run the portrait matting models on these corrected images again. Experimental results show that the under- and overexposure cause severe degradation in the performance of the portrait matting models. The experiments further show that our error correction model overcomes this degradation and causes a significant improvement in the portrait matting performance. Our contributions are as follows:
\begin{itemize}
    \item We propose an end-to-end exposure correction model by utilizing feature matching loss, perceptual loss, and multi-scale discriminator to improve the performance. 
    \item We thoroughly analyze the proposed model and show that it achieves the SOTA performance on the exposure correction benchmark dataset. Besides, the proposed model shows a good generalization capacity on four different low-light image enhancement benchmarks.
    \item We observe that under- and overexposed images cause significant degradation in the portrait matting performance. However, after the images are corrected by the proposed exposure correction model, the portrait matting models achieve significantly better performance. 
\end{itemize}

\section{Related Work}

Researchers propose various methods to enhance images for exposure correction and lighting changes. Early works focus on adjusting images with contrast-based histogram equalization methods \cite{zuiderveld1994contrast, ibrahim2007brightness, celik2011contextual, thomas2011histogram, lee2013contrast}. Then, approaches built on the Retinex theory \cite{land1977retinex} attract interest in the following years \cite{lime2,WVM,cai2017joint,RetinexNet,DeepUPE, ma2020joint}. These studies assume that images can be decomposed into reflection and illumination maps that rely on hand-picked constraints. In this case, insufficient results may occur, since obtaining accurate illumination maps to enhance image lighting is difficult in challenging scenes. In recent years, deep learning-based studies \cite{chen2018deep,yu2018deepexposure,yang2018personalized,KinD,DeepUPE,ZeroDCE,xu2020learning,afifi2020deep,ni2020towards,afifi2021learning,ouyang2021neural,yang2022learning} have gained importance in image enhancement and exposure correction literature. Until very recent studies, low light image enhancement \cite{lime2,lore2017llnet,RetinexNet,zhang2018high,loh2019getting,DeepUPE,KinD,ZeroDCE,xu2020learning,yang2020fidelity,zhu2020eemefn,DPCB,zhang2021unsupervised,zhang2021beyond,li2021low,huang2022towards} 
was the primary research direction in exposure correction works due to the limitation of collected benchmark datasets on overexposure settings. However, applying only underexposure correction does not suit real-world conditions. Therefore, overexposure errors should also be covered in more comprehensive approaches. A few works examine both over- and underexposure errors in images \cite{yang2018personalized,zhang2019dual,ma2020joint,afifi2021learning,yang2022learning,ouyang2021neural,liang2022fusion}. To the best of our knowledge, Afifi et al. \cite{afifi2021learning} is the first work that addresses the underexposure and overexposure problem at the same time by a deep learning-based method. They also propose a large-scale exposure error dataset for both  under- and overexposure cases and compare several exposure correction methods on the proposed dataset. There are more recent works that address both problems as well. In \cite{yang2022learning}, the authors propose a unified LA-Net model to investigate low-light enhancement, exposure correction, and tone mapping tasks. The proposed approach handles the light adaptation problem with so called low- and high-frequency pathways. In addition to using a single image to learn exposure correction, some studies utilize a sequence of images with multi-exposure values in order to learn enhancement \cite{cai2018learning,ma2019deep,albahar2021contrast,qu2021transmef,liang2022fusion}. Moreover, Liang et al. \cite{liang2022fusion} also exploit single exposure correction network.

In this paper, we analyze under- and overexposure with a single model to perform exposure correction. We further use adversarial loss with the feature matching loss, perceptual loss, and multi-scale discriminator to improve the quality. 
We also investigate the effect of exposure error on the portrait matting task and propose to use our model to improve the portrait matting performance by applying exposure correction.




\begin{figure*}
    \centering
    \includegraphics[width=17cm]{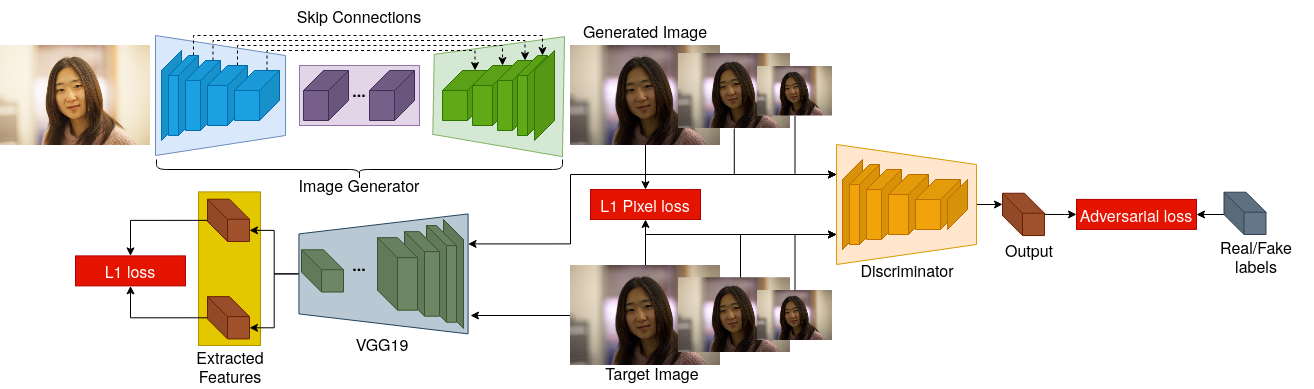}
    \caption{Proposed exposure correction method. Our generator contains encoder to embed the input image, residual block to edit the feature representation, and image decoder that is responsible to synthesize the output image. Perceptual loss, L1 pixel loss, and multi-scale discriminator lead to improve the performance.} 
    \label{fig:exposure_method}
\end{figure*}


\section{Exposure correction}

In this study, our main goal is to correct the exposure of the input image. Our model receives an image to manipulate the exposure setting in the feature space to synthesize the corrected version. There are three crucial points that we must address. 1) We must preserve the content of the image and only modify the exposure setting. 2) The input image can be underexposed or overexposed, therefore, we must learn the exposure correction for both cases without prior knowledge. 3) As we do not provide prior knowledge to the network about the exposure setting of the input image, we must also consider the scenario in which the input image has the correct exposure setting. In this case, our model must have the capacity to discover that the image has no exposure issue and must sustain the same exposure setting. Consequently, we address the exposure correction task for underexposed, overexposed, and well exposed cases by an end-to-end generative adversarial network based approach~\cite{vanillaGAN}. 

\textbf{Generator.}  Given an input image $ x \in \mathbb{R}^{W \times H \times 3} $, the objective of the image generator $ G $ is to synthesize an output image $ y' \in \mathbb{R}^{W \times H \times 3} $ that must have the correct exposure setting. Besides, $ G $ must learn not to alter the image with a correct exposure setting. The proposed $ G $ involves three different submodules; image encoder $ E_{img} $, residual block $ R $~\cite{residualblocks}, and image decoder $ D_{img} $. First, we extract image features $ \phi_{x} \in \mathbb{R}^{512 \times 16 \times 16} $ using the image encoder $ E_{img} $ and then forward it through the consecutive residual blocks in order to edit the feature representation. In the end, the altered features pass through the image decoder to generate the final output $ y' $. Moreover, we employ several residual connections between reciprocal convolutional layers of the image encoder $ E_{img} $ and decoder $ D_{img} $ to preserve the intermediate feature representation of $ E_{img} $ in the decoder $ D_{img} $. The proposed image generator $ G $ is illustrated in Figure \ref{fig:exposure_method}. Our image generator has five consecutive convolutional layers with ReLU activation function~\cite{relu1,relu2} and Instance Normalization layer~\cite{ulyanov2016instance}. Similarly, the proposed image decoder contains five consecutive transposed convolutional layer with ReLU and Instance Norm to synthesize images. 

\textbf{Discriminator.} We employ a discriminator network $ D $ to distinguish between real samples and fake samples. The proposed  $ D $ is presented in Figure~\ref{fig:exposure_method}. It contains several consecutive convolutional layers to downsample the image to produce an output representation. We employ spectral normalization~\cite{miyato2018spectral} after the convolutional layers to normalize the spectral norm of the weight matrices. This helps us to make the training more stable as well as to bound the Lipschitz norm $ \sigma(W) = 1 $. Moreover, we propose to use multiple discriminators~\cite{durugkar2016generative}. For this, we utilize three identical discriminators at three different scales~\cite{pix2pixhd}. According to our preliminary experimental analysis, this multi-scale discriminator approach improves to capture both the global structure and details of the images simultaneously and leads to synthesizing high-resolution images in better quality. The discriminator network involves five consecutive convolutional layers with spectral normalization~\cite{miyato2018spectral} and Leaky ReLU activation function~\cite{lrelu}.

\begin{figure*}
    \centering
    \includegraphics[scale=0.55]{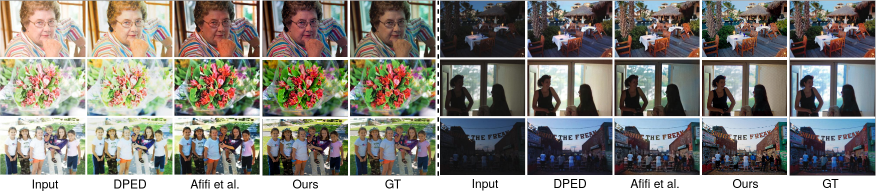}
    \caption{The results on exposure correction dataset \cite{afifi2021learning}. We take images from~\cite{afifi2021learning} and compare with our model.  } 
    \label{fig:compare3}
\end{figure*}

\definecolor{ourgreen}{HTML}{d2f0aa}
\definecolor{ouryellow}{HTML}{fcef9e}

\setlength{\tabcolsep}{0.5pt}
\begin{table*}\footnotesize
  \centering
  \begin{tabular}{@{}l|c|c|c|c|c|c|c|c|c|c|c|c|c@{}}
    \toprule
    Method & \multicolumn{2}{c}{Expert A} & \multicolumn{2}{c}{Expert B} & \multicolumn{2}{c}{Expert C} & \multicolumn{2}{c}{Expert D} & \multicolumn{2}{c}{Expert E} & \multicolumn{2}{c}{Avg} & PI \\
     & PSNR & SSIM & PSNR & SSIM & PSNR & SSIM & PSNR & SSIM & PSNR & SSIM & PSNR & SSIM & \\
    \midrule
    WVM~\cite{WVM} & 12.355 & 0.624 & 13.147 & 0.656 & 12.748 & 0.645 & 14.059 & 0.669 & 15.207 & 0.690 & 13.503 & 0.657 & 2.342 \\
    LIME*~\cite{lime1,lime2} & 09.627 & 0.549 & 10.096 & 0.569 & 9.875 & 0.570 & 10.936 & 0.597 & 11.903 & 0.626 & 10.487 & 0.582 & 2.412 \\
    HDR CNN w/ PS~\cite{HDRCNN} & 14.804 & 0.651 & 15.622 & 0.689 & 15.348 & 0.670 & 16.583 & 0.685 & 18.022 & 0.703 & 16.076 & 0.680 & 2.248 \\
    DPED (iPhone)~\cite{DPED} & 12.680 & 0.562 & 13.422 & 0.586 & 13.135 & 0.581 & 14.477 & 0.596 & 15.702 & 0.630 & 13.883 & 0.591 & 2.909 \\
    DPED (BlackBerry)~\cite{DPED} & 15.170 & 0.621 & 16.193 & 0.691 & 15.781 & 0.642 & 17.042 & 0.677 & 18.035 & 0.678 & 16.444 & 0.662 & 2.518 \\
    DPE (HDR)~\cite{chen2018deep} & 14.399 & 0.572 & 15.219 & 0.573 & 15.091 & 0.593 & 15.692 & 0.581 & 16.640 & 0.626 & 15.408 & 0.589 & 2.417 \\
    DPE (S-FiveK)~\cite{chen2018deep} & 14.786 & 0.638 & 15.519 & 0.649 & 15.625 & 0.668 & 16.586 & 0.664 & 17.661 & 0.684 & 16.035 & 0.661 & 2.621 \\
    RetinexNet*~\cite{RetinexNet} & 10.149 & 0.570 & 10.880 & 0.586 & 10.471 & 0.595 & 11.498 & 0.613 & 12.295 & 0.635 & 11.059 & 0.600 & 2.933 \\
    Deep UPE*~\cite{DeepUPE} & 10.047 & 0.532 & 10.462 & 0.568 & 10.307 & 0.557 & 11.583 & 0.591 & 12.639 & 0.619 & 11.008 & 0.573 & 2.428 \\
    Zero-DCE~\cite{ZeroDCE} & 10.116 & 0.503 & 10.767 & 0.502 & 10.395 & 0.514 & 11.471 & 0.522 & 12.354 & 0.557 & 11.021 & 0.519 & 2.774 \\
    Afifi et al. w/o $ L_{adv} $~\cite{afifi2021learning} & \colorbox{ouryellow}{18.976} & \colorbox{ouryellow}{0.743} & \colorbox{ouryellow}{19.767} & \colorbox{ouryellow}{0.731} & \colorbox{ouryellow}{19.980} & \colorbox{ouryellow}{0.768} & \colorbox{ouryellow}{18.966} & \colorbox{ouryellow}{0.716} & \colorbox{ouryellow}{19.056} & \colorbox{ouryellow}{0.727} & \colorbox{ouryellow}{19.349} & \colorbox{ouryellow}{0.737} & 2.189 \\
    Afifi et al. w/ $ L_{adv} $ ~\cite{afifi2021learning} & 18.874 & 0.738 & 19.569 & 0.718 & 19.788 & 0.760 & 18.823 & 0.705 & 18.936 & 0.719 & 19.198 & 0.728 & \colorbox{ouryellow}{2.183} \\
    Ours & \colorbox{ourgreen}{20.475} & \colorbox{ourgreen}{0.862} & \colorbox{ourgreen}{21.833} & \colorbox{ourgreen}{0.891} & \colorbox{ourgreen}{22.438} & \colorbox{ourgreen}{0.901} & \colorbox{ourgreen}{20.127} & \colorbox{ourgreen}{0.874} & \colorbox{ourgreen}{20.062} & \colorbox{ourgreen}{0.881} & \colorbox{ourgreen}{20.987} & \colorbox{ourgreen}{0.881} & \colorbox{ourgreen}{2.158} \\
    \hline
    WVM~\cite{WVM} & 17.686 & 0.728 & 19.787 & \colorbox{ouryellow}{0.764} & 18.670 & 0.728 & 18.568 & 0.729 & 18.362 & 0.724 & 18.615 & 0.735 & 2.525 \\
    LIME*~\cite{lime1,lime2} & 13.444 & 0.653 & 14.426 & 0.672 & 13.980 & 0.663 & 15.190 & 0.673 & 16.177 & 0.694 & 14.643 & 0.671 & 2.462 \\
    HDR CNN w/ PS~\cite{HDRCNN} & 17.324 & 0.692 & 18.992 & 0.714 & 18.047 & 0.696 & 18.377 & 0.689 & \colorbox{ouryellow}{19.593} & 0.701 & 18.467 & 0.698 & \colorbox{ourgreen}{2.294} \\
    DPED (iPhone)~\cite{DPED} & 18.814 & 0.680 & 21.129 & 0.712 & 20.064 & 0.683 & \colorbox{ouryellow}{19.711} & 0.675 & 19.574 & 0.676 & 19.858 & 0.685 & 2.894 \\
    DPED (BlackBerry)~\cite{DPED} & 19.519 & 0.673 & \colorbox{ourgreen}{22.333} & 0.745 & 20.342 & 0.669 & 19.611 & 0.683 & 18.489 & 0.653 & \colorbox{ouryellow}{20.059} & 0.685 & 2.633 \\
    DPE (HDR)~\cite{chen2018deep} & 17.625 & 0.675 & 18.542 & 0.705 & 18.127 & 0.677 & 16.831 & 0.665 & 15.891 & 0.643 & 17.403 & 0.673 & \colorbox{ouryellow}{2.340} \\
    DPE (S-FiveK)~\cite{chen2018deep} & \colorbox{ouryellow}{20.153} & 0.738 & 20.973 & 0.697 & \colorbox{ouryellow}{20.915} & 0.738 & 19.050 & 0.688 & 17.510 & 0.648 & 19.720 & 0.702 & 2.564 \\
    RetinexNet*~\cite{RetinexNet} & 11.676 & 0.607 & 12.711 & 0.611 & 12.132 & 0.621 & 12.720 & 0.618 & 13.233 & 0.637 & 12.494 & 0.619 & 3.362 \\
    Deep UPE*~\cite{DeepUPE} & 17.832 & 0.728 & 19.059 & 0.754 & 18.763 & 0.745 & 19.641 & \colorbox{ouryellow}{0.737} & \colorbox{ourgreen}{20.237} & \colorbox{ouryellow}{0.740} & 19.106 & 0.741 & 2.371 \\
    Zero-DCE~\cite{ZeroDCE} & 13.935 & 0.585 & 15.239 & 0.593 & 14.552 & 0.589 & 15.202 & 0.587 & 15.893 & 0.614 & 14.964 & 0.593 & 3.001 \\
    Afifi et al. w/o $ L_{adv} $ ~\cite{afifi2021learning} & 19.432 & 0.750 & 20.590 & 0.739 & 20.542 & \colorbox{ouryellow}{0.770} & 18.989 & 0.723 & 18.874 & 0.727 & 19.685 & \colorbox{ouryellow}{0.742} & 2.344 \\
    Afifi et al. w/ $ L_{adv} $ ~\cite{afifi2021learning} & 19.475 & \colorbox{ouryellow}{0.751} & 20.546 & 0.730 & 20.518 & 0.768 & 18.935 & 0.715 & 18.756 & 0.719 & 19.646 & 0.737 & 2.342  \\
    Ours & \colorbox{ourgreen}{20.397} & \colorbox{ourgreen}{0.858} & \colorbox{ouryellow}{21.683} & \colorbox{ourgreen}{0.883} & \colorbox{ourgreen}{22.175} & \colorbox{ourgreen}{0.893} & \colorbox{ourgreen}{19.771} & \colorbox{ourgreen}{0.865} & 19.508 & \colorbox{ourgreen}{0.867} & \colorbox{ourgreen}{20.706} & \colorbox{ourgreen}{0.873} & 2.375 \\
    \hline
    WVM~\cite{WVM} & 14.488 & 0.665 & 15.803 & 0.699 & 15.117 & 0.678 & 15.863 & 0.693 & 16.469 & 0.704 & 15.548 & 0.688 & 2.415 \\
    LIME*~\cite{lime1,lime2} & 11.154 & 0.591 & 11.828 & 0.610 & 11.517 & 0.607 & 12.638 & 0.628 & 13.613 & 0.653 & 12.150 & 0.618 & 2.432 \\
    HDR CNN w/ PS~\cite{HDRCNN} & 15.812 & 0.667 & 16.970 & 0.699 & 16.428 & 0.681 & 17.301 & 0.687 & 18.650 & 0.702 & 17.032 & 0.687 & 2.267 \\
    DPED (iPhone)~\cite{DPED} & 15.134 & 0.609 & 16.505 & 0.636 & 15.907 & 0.622 & 16.571 & 0.627 & 17.251 & 0.649 & 16.274 & 0.629 & 2.903 \\
    DPED (BlackBerry)~\cite{DPED} & 16.910 & 0.642 & 18.649 & 0.713 & 17.606 & 0.653 & 18.070 & 0.679 & 18.217 & 0.668 & 17.890 & 0.671 & 2.564 \\
    DPE (HDR)~\cite{chen2018deep} & 15.690 & 0.614 & 16.548 & 0.626 & 16.305 & 0.626 & 16.147 & 0.615 & 16.341 & 0.633 & 16.206 & 0.623 & 2.417 \\
    DPE (S-FiveK)~\cite{chen2018deep} & 16.933 & 0.678 & 17.701 & 0.668 & 17.741 & 0.696 & 17.572 & 0.674 & 17.601 & 0.670 & 17.510 & 0.677 & 2.621 \\
    RetinexNet*~\cite{RetinexNet} & 10.759 & 0.585 & 11.613 & 0.596 & 11.135 & 0.605 & 11.987 & 0.615 & 12.671 & 0.636 & 11.633 & 0.607 & 3.105 \\
    Deep UPE*~\cite{DeepUPE} & 13.161 & 0.610 & 13.901 & 0.642 & 13.689 & 0.632 & 14.806 & 0.649 & 15.678 & 0.667 & 14.247 & 0.640 & 2.405 \\
    Zero-DCE~\cite{ZeroDCE} & 11.643 & 0.536 & 12.555 & 0.539 & 12.058 & 0.544 & 12.964 & 0.548 & 13.769 & 0.580 & 12.597 & 0.549 & 2.865 \\
    Afifi et al. w/o $ L_{adv} $ ~\cite{afifi2021learning} & \colorbox{ouryellow}{19.158} & \colorbox{ouryellow}{0.746} & \colorbox{ouryellow}{20.096} & \colorbox{ouryellow}{0.734} & \colorbox{ouryellow}{20.205} & \colorbox{ouryellow}{0.769} & \colorbox{ouryellow}{18.975} & \colorbox{ouryellow}{0.719} & \colorbox{ouryellow}{18.983} & \colorbox{ouryellow}{0.727} & \colorbox{ouryellow}{19.483} & \colorbox{ouryellow}{0.739} & 2.251 \\
    Afifi et al. w/ $ L_{adv} $ ~\cite{afifi2021learning} & 19.114 & 0.743 & 19.960 & 0.723 & 20.080 & 0.763 & 18.868 & 0.709 & 18.864 & 0.719 & 19.377 & 0.731 & \colorbox{ouryellow}{2.247} \\
    Ours & \colorbox{ourgreen}{20.443} & \colorbox{ourgreen}{0.860} & \colorbox{ourgreen}{21.773} & \colorbox{ourgreen}{0.887} & \colorbox{ourgreen}{22.332} & \colorbox{ourgreen}{0.897} & \colorbox{ourgreen}{19.984} & \colorbox{ourgreen}{0.870} & \colorbox{ourgreen}{19.840} & \colorbox{ourgreen}{0.875} & \colorbox{ourgreen}{20.874} & \colorbox{ourgreen}{0.877} & \colorbox{ourgreen}{2.244} \\ 
    \hline
  \end{tabular}
  \caption{Exposure correction results on exposure dataset\cite{afifi2021learning}. We highlight the best results with green and the second-best results with yellow. $ * $ shows the methods that were trained for only the underexposure case. We divide the table into three subgroups: 1) well- and overexposure (3543 images), 2) underexposure (2362 images), 3) altogether case (5905 images).}
  \label{tab:exposure1}
\end{table*}
\setlength{\tabcolsep}{1.4pt}

\begin{figure*}
    \centering
    \includegraphics[scale=0.35]{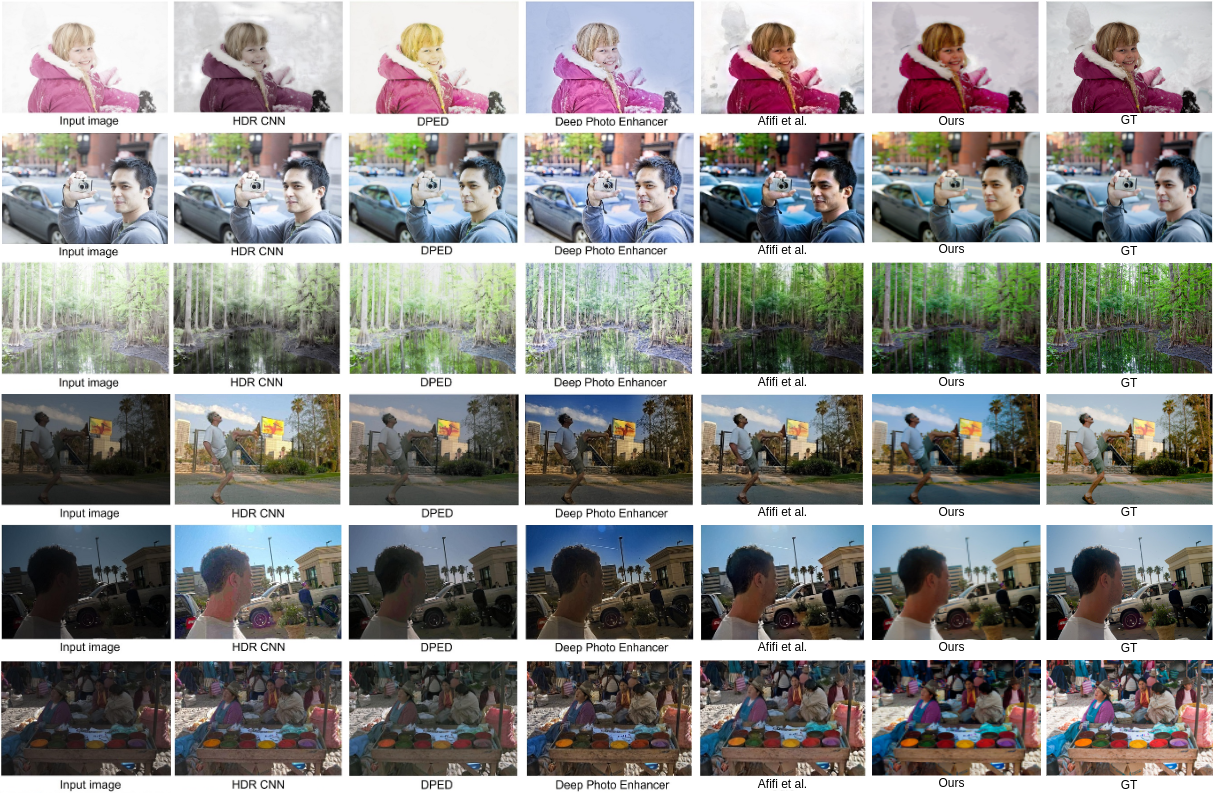}
    \caption{We take outputs of other models from~\cite{afifi2021learning} and compare with our results and ground truth images from \textit{expert A}. }
    \label{fig:compare1}
\end{figure*}

\subsection{Learning procedure}

The outputs of our multi-scale discriminator are $ \mathbb{R}^{N \times N \times 1}, N=16,32,64 $. 
We use these feature representations 
of real and fake samples in the adversarial loss. For this, we create real and fake matrices with one and zero values as the discriminator output. 
and we calculate MSE between them. 
Besides, we utilize improved adversarial loss~\cite{pix2pixhd} to stabilize the training by extracting features from different levels of the discriminator and calculate L1 distance to perform feature matching at different scales. 
Additionally, we 
benefit from the perceptual loss~\cite{johnson2016perceptual} by extracting features from the real and fake samples using the VGG-19 network~\cite{vgg} to compare the L1 distance in the feature space. The corresponding loss function is shown in Equation~\ref{equation:perceptual}.

\begin{equation}
    L_{per} =  \sum_{i=1}^{5} (c_i ||(L^{i}(y') - L^{i}(y)||_1
    \label{equation:perceptual}
\end{equation}

We use five different layers to extract features and calculate the loss~\cite{johnson2016perceptual}. Furthermore, we employ L1 distance loss in the pixel space to directly compare the generated image and the ground truth image: $    L_{pixel} = ||y' - y||_1 $. The overall loss is presented in Equation~\ref{equation:overall}.

\begin{equation}
    L = L_{GAN}(G, D) + \lambda L_{pixel}(G) + \beta L_{per}(G)
    \label{equation:overall}
\end{equation}

Utilized coefficients $\lambda$ and $\beta$ adjust the effect of each loss over the final loss. We empirically find the best values 
on the validation set of 
~\cite{afifi2021learning} as 0.5 and 1, respectively.


\section{Experimental Result}

\subsection{Datasets}
\textbf{Exposure correction dataset.} We train our model on a novel large-scale exposure correction dataset~\cite{afifi2021learning} which has underexposed, overexposed, well-exposed, and ground truth images. We follow the same setup in~\cite{afifi2021learning}. The training set contains 17675 images, while the validation set has 750 images. 
Lastly, the test set consists of 5905 images with five different versions: well-exposed, underexposed with -1 and -1.5 EVs, and overexposed with +1 and +1.5 EVs. 

\textbf{Portrait matting datasets.} We also investigate the effect of under- and overexposed images over the portrait matting performance. For this, we 
choose four different real-world portrait matting datasets. Since we manipulate the exposure setting of the images, we need to have real-world images to investigate the exposure accurately. In order to perform this analysis, we utilize PPM-100~\cite{modnet}, P3M500~\cite{p3m500}, RWP636~\cite{mgm}, and AIM500~\cite{aim500} datasets. These datasets 
contain 100, 500, 636, and 500 images, respectively. Please note that AIM500~\cite{aim500} is proposed for general image matting, therefore, we only select 
portrait 
images from this dataset. In the end, we have 100 images for this dataset. 
We use Adobe Photoshop Lightroom to digitally manipulate the exposure setting of these datasets to obtain under- and overexposed images by utilizing -1.5, -1, +1, and +1.5 values. 
For further investigation, we also utilize -2, -2.5, +2, and +2.5 EVs to change the exposure setting more. With this additional manipulation, we both examine the effect of severe exposure errors and the performance of the exposure correction model beyond the used cases in the training.






\subsection{Evaluation}

In order to assess 
the exposure correction performance, we follow the literature and use three evaluation metrics: peak signal-to-noise ratio (PSNR), structural similarity index measure (SSIM)~\cite{ssim}, and perceptual index (PI)~\cite{PI_metric} which is ground truth-free evaluation method and is the combination of Ma~\cite{ma} and NIQE~\cite{niqe} metrics.

In order to evaluate the effect of the exposure setting over the portrait matting, we decide to use three state-of-the-art matting methods that do not need to have an additional input to produce an alpha matte. We utilize MODNet~\cite{modnet}, MGMatting (MGM)~\cite{mgm}, and GFM~\cite{GFM}. We run these models with well-, under-, and overexposed images to generate alpha matte. Then, we repeat the same experiments with the images in which their exposure setting is corrected by our exposure correction model. To evaluate the outputs, we use two commonly used metrics in the matting literature: mean squared error (MSE) and mean absolute error (MAE).

\subsection{Exposure Correction Results}

In Table~\ref{tab:exposure1}, we present the results on exposure correction dataset~\cite{afifi2021learning}. We follow the same analysis in \cite{afifi2021learning} and split the table into three separate groups. In the first group, we show the results for well- and overexposed images that are obtained with $ 0, +1, +1.5 $ EVs. This test setup contains 3543 images. In the second one, we demonstrate the results for underexposure cases, with EVs of $ -1$ and $ -1.5 $. This test setup involves 2362 images. In the last group, we present the 
results for all cases and this includes all 5905 images. 

The corresponding dataset has five different ground truth images for each input, namely \textit{expert A, expert B, expert C, expert D, expert E}. Therefore, we provide the results for all of these five experts separately. In the end, we also calculate the average results 
over these expert sets. Please note that we acquire all these results, except \textit{Ours}, from the corresponding paper~\cite{afifi2021learning}. 
Experimental results show 
that our proposed method outperforms 
all other methods for all scenarios, experts, and metrics. Only two exceptions are DPED (BlackBerry) method that achieves slightly better PSNR result for underexposure scenario for Expert B ground truth set and Deep UPE method which obtains slightly higher PSNR score for Expert E case. 
Please note that Deep UPE was proposed for only underexposure problem. Because of that it is more likely for it to achieve better performance for the underexposure scenario in theory. However, our proposed method surpasses DPED (BlackBerry) and Deep UPE models in terms of the SSIM metric for Expert B and Expert E. Besides, for the rest of the experts, our method surpasses all other methods in both metrics. In summary, our method has the best average scores as can be seen from the last row of Table~\ref{tab:exposure1}, since the proposed system has more complex and effective architecture with residual connections for conditional image generation. Moreover, utilized loss functions and multi-scale discriminator improve the quality of the generated image and make the learning procedure more stable. With the help of all these components, the proposed system is able to adjust the exposure setting without modifying the content.  
Please note that, Deep UPE, RetinexNet, and LIME models are specifically developed for the underexposure case only. However, although our proposed method is not specifically designed for underexposure correction, it still outperforms these models on average for the underexposure case. 
In Table~\ref{tab:exposure2}, we present the results of recent works on exposure correction dataset~\cite{afifi2021learning}. According to the table, our proposed model achieves state-of-the-art results by surpassing all other methods on three different metrics.

\begin{figure}
    \centering
    \includegraphics[scale=0.25]{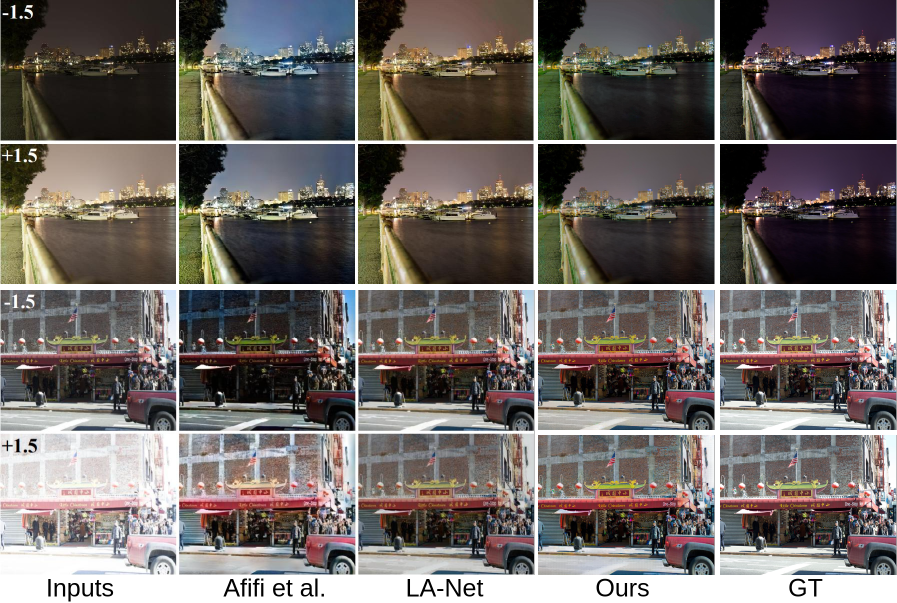}
    \caption{Visual comparisons of sample images with under- and overexposure conditions. We compare our results with Afifi et al.~\cite{afifi2021learning}, LA-Net~\cite{yang2022learning}, and ground truth images. }
    \label{fig:compare4}
\end{figure}

\setlength{\tabcolsep}{4pt}
\begin{table}\footnotesize
  \centering
  \begin{tabular}{@{}l|ccc@{}}
    \toprule
    Method & PSNR $ \uparrow $ & SSIM $ \uparrow $ & PI $ \downarrow $ \\
    \midrule
    Afifi et al. w/o $ L_{adv} $ ~\cite{afifi2021learning} & 19.48 & 0.73 & 2.251 \\
    Afifi et al. w/ $ L_{adv} $ ~\cite{afifi2021learning} & 19.37 & 0.73 & 2.247 \\
    LA-Net~\cite{yang2022learning} & 20.70 & 0.81 & 2.353 \\
    FCN SEC~\cite{liang2022fusion} & 19.71 & 0.80 & - \\
    FCN MEF~\cite{liang2022fusion} & 20.81 & 0.84 & - \\
    Ours & \textbf{20.87} & \textbf{0.87} & \textbf{2.244} \\
    \bottomrule
  \end{tabular}
  \caption{Summary of exposure correction results on \cite{afifi2021learning}. We achieve the state-of-the-art result on all three metrics. }
  \label{tab:exposure2}
\end{table}
\setlength{\tabcolsep}{1.4pt}


In Figure~\ref{fig:compare3}, \ref{fig:compare1}, \ref{fig:compare4}, we present different test images, our results, ground truth images, and the results of other models. We obtain the corresponding outputs for other models from their papers in order to compare with our results. 
As one can see from Figure~\ref{fig:compare3} and Figure~\ref{fig:compare1}, our results are closer to the ground truth images than the other methods. Besides, our results have better quality in the details and provide a more similar color distribution with the ground truth images. In Figure~\ref{fig:compare4}, we provide four different cases and our model can accurately correct the exposure problem, although all three models have a problem in the color distribution of the first image due to the challenging lighting condition.


\begin{figure}
\footnotesize
\centering
\includegraphics[scale=0.20]{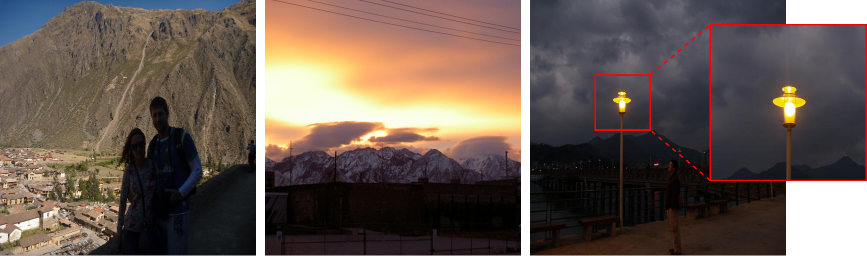}\\
Inputs \\
\includegraphics[scale=0.20]{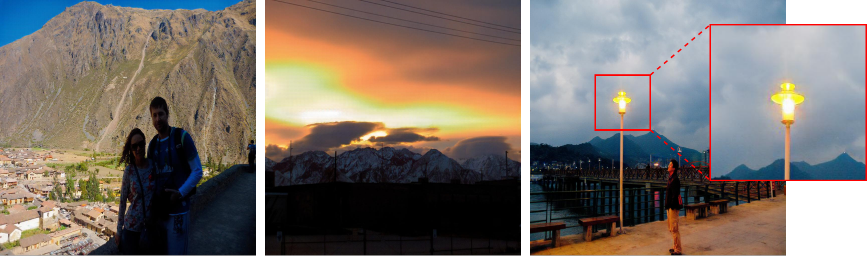} \\
Outputs
\caption{Failure cases. 
The model may fail when the image includes locally under- and overexposed parts. Output may also have artifacts around light sources, despite well corrected exposure.  }
\label{fig:limitations}
\end{figure}


\setlength{\tabcolsep}{1.4pt}
\begin{table}\footnotesize
  \centering
  \begin{tabular}{@{}l|ccccc@{}}
    \toprule
    Method & LIME~\cite{lime2} & NPE~\cite{NPE} & VV~\cite{VV} & DICM~\cite{DICM} & Avg. \\
    \midrule
    NPE~\cite{NPE} & 3.91 & 3.95 & 2.52 & 3.76 & 3.54 \\
    LIME~\cite{lime2} & 4.16 & 4.26 & 2.49 & 3.85 & 3.69 \\
    WVM~\cite{WVM} & 3.79 & 3.99 & 2.85 & 3.90 & 3.63 \\
    RNet~\cite{RetinexNet} & 4.42 & 4.49 & 2.60 & 4.20 & 3.93 \\
    KinD~\cite{KinD} & 3.72 & 3.88 & - & - & 3.80 \\
    EGAN~\cite{EGAN} & 3.72 & 4.11 & 2.58 &- & 3.50 \\
    DBCP~\cite{DPCB} & 3.78 & 3.18 & - & 3.57 & 3.48 \\
    Afifi et al.~\cite{afifi2021learning} & 3.76 & 3.18 & \textbf{2.28} & \textbf{2.50} & \textbf{2.93} \\
    Ours & \textbf{3.12} & \textbf{2.90} & 2.98 & 3.02 & 3.01 \\
    \bottomrule
  \end{tabular}
  \caption{Results on four low-light image datasets. We follow the literature and provide NIQE metric on these datasets. Lower results indicate better performance.} 
  \label{tab:low_light}
\end{table}
\setlength{\tabcolsep}{1.4pt}

\setlength{\tabcolsep}{1pt}
\begin{table*}\footnotesize
  \centering
  \begin{tabular}{@{}l|c|cc|cc|cc|cc|cc|cc|cc|cc|cc|cc@{}}
    \toprule
    Data & Method & \multicolumn{2}{c}{-2.5} & \multicolumn{2}{c}{-2} & \multicolumn{2}{c}{-1.5} & \multicolumn{2}{c}{-1} & \multicolumn{2}{c}{0} & \multicolumn{2}{c}{1} & \multicolumn{2}{c}{1.5} & \multicolumn{2}{c}{2} & \multicolumn{2}{c}{2.5} & \multicolumn{2}{c}{Avg} \\
     &  & MSE & MAE & MSE & MAE & MSE & MAE & MSE & MAE & MSE & MAE & MSE & MAE & MSE & MAE & MSE & MAE & MSE & MAE & MSE & MAE \\
    \midrule
    PPM-100 E & MODNet & 10.1 & 16.7 & 8.6 & 15.0 & 7.3 & 13.8 & \textbf{6.1} & \textbf{12.1} & \textbf{5.8} & \textbf{11.5} & \textbf{5.6} & 11.5 & 6.6 & 12.6 & 6.7 & 12.7 & 10 & 16.4 & 7.42 & 13.59 \\
    
    PPM-100 C & MODNet & \textbf{7.8} & \textbf{13.4} & \textbf{6.9} & \textbf{12.5} & \textbf{6.6} & \textbf{12.2} & 6.6 & 12.2 & 6.2 & 11.7 & 6.0 & \textbf{11.1} & \textbf{6.0} & \textbf{11.2} & \textbf{5.7} & \textbf{11.7} & \textbf{6.0} & \textbf{12.5} & \textbf{6.42} & \textbf{12.05} \\ \hline
    P3M500 E & MODNet & 9.9 & 17.3 & 8.7 & 16.0 & 8.1 & 15.0 & 7.8 & 14.7 & 7.5 & 14.3 & 10.2 & 16.9 & 11.8 & 18.5 & 14.5 & 21.6 & 19.7 & 27.5 & 10.91 & 17.97 \\
    
    P3M500 C & MODNet & \textbf{7.8} & \textbf{14} & \textbf{7.0} & \textbf{13.1} & \textbf{6.9} & \textbf{13.1} & \textbf{6.6} & \textbf{12.7} & \textbf{6.1} & \textbf{11.9} & \textbf{6.2} & \textbf{12.2} & \textbf{6.5} & \textbf{12.5} & \textbf{7.4} & \textbf{13.4} & \textbf{10.0} & \textbf{16.1} & \textbf{7.26} & \textbf{13.22} \\ 
    \hline
    RWP636 E & MODNet & 27.2 & 41.8 & 24.8 & 39.3 & 22.5 & 36.8 & 22.1 & 35.4 & 24.3 & 38.0 & 27.6 & 41.7 & 33.9 & 48.4 & 40.7 & 55.5 & 50.7 & 66.0 & 30.42 & 44.76 \\
    
    RWP636 C & MODNet & \textbf{22.4} & \textbf{35.4} & \textbf{21.9} & \textbf{34.9} & \textbf{21.4} & \textbf{34.4} & \textbf{20.5} & \textbf{33.4} & \textbf{19.9} & \textbf{32.1} & \textbf{20.3} & \textbf{33.1} & \textbf{20.4} & \textbf{33.4} & \textbf{21.5} & \textbf{34.5} & \textbf{24.5} & \textbf{37.5} & \textbf{21.42} & \textbf{34.30} \\ 
    \hline
    AIM500 E & MODNet & 11.5 & 18.2 & 10.8 & 17.4 & \textbf{10.1} & \textbf{16.5} & 10.0 & 16.3 & 10.0 & 16.3 & 12.1 & 18.1 & 13.1 & 19.3 & 15.3 & 22.0 & 19.4 & 26.7 & 12.47 & 18.97 \\
    
    AIM500 C & MODNet & \textbf{10.9} & \textbf{17.0} & \textbf{10.6} & \textbf{16.8} & 10.6 & 16.7 & \textbf{9.9} & \textbf{15.9} & \textbf{9.8} & \textbf{15.8} & \textbf{10.4} & \textbf{16.2} & \textbf{10.7} & \textbf{16.6} & \textbf{10.4} & \textbf{16.3} & \textbf{11.7} & \textbf{17.7} & \textbf{10.55} & \textbf{16.55} \\
    \hline 
    PPM-100 E & MGM & 1.6 & 6.3 & \textbf{1.3} & 5.9 & \textbf{1.3} & \textbf{5.6} & \textbf{1.2} & \textbf{5.4} & \textbf{1.2} & \textbf{5.3} & 1.4 & 5.7 & 1.6 & 6.1 & 1.8 & 6.5 & 2.0 & 7.0 & 1.49 & 5.98 \\
    
    PPM-100 C & MGM & \textbf{1.3} & \textbf{5.8} & \textbf{1.3} & \textbf{5.7} & \textbf{1.3} & \textbf{5.6} & \textbf{1.2} & \textbf{5.4} & 1.3 & 5.7 & \textbf{1.3} & \textbf{5.7} & \textbf{1.3} & \textbf{5.7} & \textbf{1.3} & \textbf{5.8} & \textbf{1.4} & \textbf{6.0} & \textbf{1.30} & \textbf{5.71} \\ 
    \hline
    P3M500 E & MGM & 4.3 & 10.5 & 4.3 & 10.3 & 3.8 & 9.4 & \textbf{3.6} & \textbf{8.9} & \textbf{3.5} & \textbf{8.3} & \textbf{3.6} & \textbf{8.6} & 4.1 & 9.6 & 4.9 & 11.1 & 6.5 & 14.2 & 4.29 & 10.10 \\
    
    P3M500 C & MGM & \textbf{3.7} & \textbf{9.1} & \textbf{3.7} & \textbf{9.1} & \textbf{3.6} & \textbf{9.0} & \textbf{3.6} & \textbf{8.9} & \textbf{3.5} & 8.7 & \textbf{3.6} & 8.8 & \textbf{3.6} & \textbf{9.0} & \textbf{3.9} & \textbf{9.6} & \textbf{5.4} & \textbf{12.8} & \textbf{3.84} & \textbf{9.44} \\ 
    \hline
    RWP636 E & MGM & 9.2 & 21.0 & 9.1 & 20.7 & \textbf{8.9} & 20.5 & \textbf{8.8} & 20.2 & \textbf{8.7} & \textbf{19.7} & 9.2 & 20.2 & 9.6 & 21.0 & 10.2 & 22.1 & 11.3 & 24.4 & 9.44 & 21.08 \\
    
    RWP636 C & MGM & \textbf{9.0} & \textbf{20.2} & \textbf{9.0} & \textbf{20.2} & 9.0 & \textbf{20.2} & \textbf{8.8} & \textbf{20.1} & \textbf{8.7} & \textbf{19.7} & \textbf{8.9} & \textbf{20.1} & \textbf{9.0} & \textbf{20.2} & \textbf{9.1} & \textbf{20.5} & \textbf{9.6} & \textbf{21.3} & \textbf{9.01} & \textbf{20.27} \\ 
    \hline
    AIM500 E & MGM & 4.4 & 10.9 & 4.2 & 10.3 & 4.1 & 9.8 & 4.3 & 9.7 & 5.8 & 10.6 & 5.0 & 10.0 & 5.6 & 11.2 & 6.6 & 13.2 & 7.8 & 16.1 & 5.31 & 11.31 \\
    
    AIM500 C & MGM & \textbf{3.9} & \textbf{9.3} & \textbf{3.6} & \textbf{9.0} & \textbf{3.7} & \textbf{9.0} & \textbf{3.8} & \textbf{9.0} & \textbf{3.7} & \textbf{8.6} & \textbf{4.2} & \textbf{9.1} & \textbf{3.9} & \textbf{9.1} & \textbf{3.4} & \textbf{8.5} & \textbf{4.3} & \textbf{10.0} & \textbf{3.83} & \textbf{9.06} \\
    \hline
    PPM-100 E & GFM & 14.3 & 14.9 & 14.0 & 14.5 & 14.0 & 14.5 & 13.9 & 14.3 & \textbf{13.6} & \textbf{14.1} & \textbf{13.3} & \textbf{13.8} & 13.5 & 14.0 & 14.2 & 14.7 & 15.2 & 15.7 & 14.00 & 14.50 \\
    
    PPM-100 C & GFM & \textbf{13.4} & \textbf{13.9} & \textbf{13.2} & \textbf{13.7} & \textbf{13.5} & \textbf{13.9} & \textbf{13.6} & \textbf{14.0} & 13.7 & \textbf{14.1} & 13.7 & 14.2 & \textbf{13.4} & \textbf{13.9} & \textbf{12.7} & \textbf{13.2} & \textbf{13.0} & \textbf{13.5} & \textbf{13.35} & \textbf{13.82} \\ 
    \hline
    P3M500 E & GFM & \textbf{8.4} & \textbf{9.0} & \textbf{8.4} & \textbf{9.0} & \textbf{8.5} & \textbf{9.1} & \textbf{8.6} & \textbf{9.1} & \textbf{9.4} & \textbf{9.9} & \textbf{11.0} & \textbf{11.5} & \textbf{12.1} & \textbf{12.6} & \textbf{13.2} & \textbf{13.7} & \textbf{14.7} & \textbf{15.2} & \textbf{10.48} & \textbf{11.01} \\
    
    P3M500 C & GFM & 13.1 & 13.6 & 13.3 & 13.8 & 13.4 & 14.0 & 13.6 & 14.1 & 14.0 & 14.5 & 14.5 & 15.0 & 14.7 & 15.2 & 14.9 & 15.4 & 15.2 & 15.7 & 14.07 & 14.58 \\ 
    \hline
    RWP636 E & GFM & \textbf{13.7} & \textbf{14.9} & 14.1 & 15.3 & 14.4 & 16.0 & 13.2 & 14.7 & \textbf{12.9} & \textbf{14.1} & 17.7 & 18.7 & 19.2 & 20.2 & 21.0 & 22.2 & 23.8 & 24.8 & 16.67 & 17.88 \\
    
    RWP636 C & GFM & \textbf{13.7} & \textbf{14.9} & \textbf{13.9} & \textbf{15.1} & \textbf{14.1} & \textbf{15.8} & \textbf{13.0} & \textbf{14.5} & \textbf{12.9} & \textbf{14.1} & \textbf{15.6} & \textbf{16.7} & \textbf{15.7} & \textbf{16.9} & \textbf{15.9} & \textbf{17.0} & \textbf{16.1} & \textbf{17.3} & \textbf{14.63} & \textbf{15.81} \\ 
    \hline
    AIM500 E & GFM & \textbf{6.8} & \textbf{7.4} & \textbf{6.6} & \textbf{7.1} & \textbf{6.5} & \textbf{7.0} & \textbf{6.5} & \textbf{7.0} & \textbf{6.4} & \textbf{6.9} & \textbf{8.4} & \textbf{8.9} & \textbf{9.3} & \textbf{9.8} & \textbf{10.2} & \textbf{10.7} & 11.8 & 12.4 & \textbf{8.05} & \textbf{8.58} \\
    
    AIM500 C & GFM & 10.3 & 10.8 & 10.4 & 10.9 & 11.0 & 11.5 & 10.9 & 11.4 & 11.1 & 11.5 & 10.9 & 11.4 & 11.2 & 11.7 & 11.0 & 11.5 & \textbf{11.5} & \textbf{12.0} & 10.92 & 11.41 \\
    \hline
  \end{tabular}
  \caption{Analysis of exposure correction in portrait matting task. White rows with \textit{E} symbol indicates the manipulated images with the corresponding EV and \textit{C} symbol in the gray rows states the corrected version of this corresponding image. }
  \label{tab:portrait_matting}
\end{table*}
\setlength{\tabcolsep}{1pt}

\textbf{Generalization}. We further perform additional tests on the four popular low-light image enhancement datasets in the literature, namely LIME~\cite{lime2}, NPE~\cite{NPE}, VV~\cite{VV}, and DICM~\cite{DICM}, and compare our method with the existing methods in the literature. We demonstrate this analysis in Table~\ref{tab:low_light}. 
Some of these datasets also contain challenging images that have under- and overexposed regions at the same time. The proposed method is found to be superior compared to the existing methods on LIME and NPE datasets, and achieve comparable results on VV and DICM datasets. This outcome indicates that our model is well-generalized and is able to work on unseen datasets. 

\textbf{Limitations.} In Figure \ref{fig:limitations}, some failure examples from the low-light image enhancement datasets are shown. 
In the first column, the model can correct the exposure of the background scene, however, bottom right part still suffers from the underexposure problem. This should be investigated in detail in the future work, since containing this kind of 
local exposure problems may sometimes be challenging. Although our proposed model can accurately adjust exposure values of the images, some cases could be difficult to handle and cause artifacts around the light sources due to the intense light changes. This can be seen from the second and third examples in Figure~\ref{fig:limitations}.

\begin{figure*}
    \centering
    \includegraphics[scale=0.19]{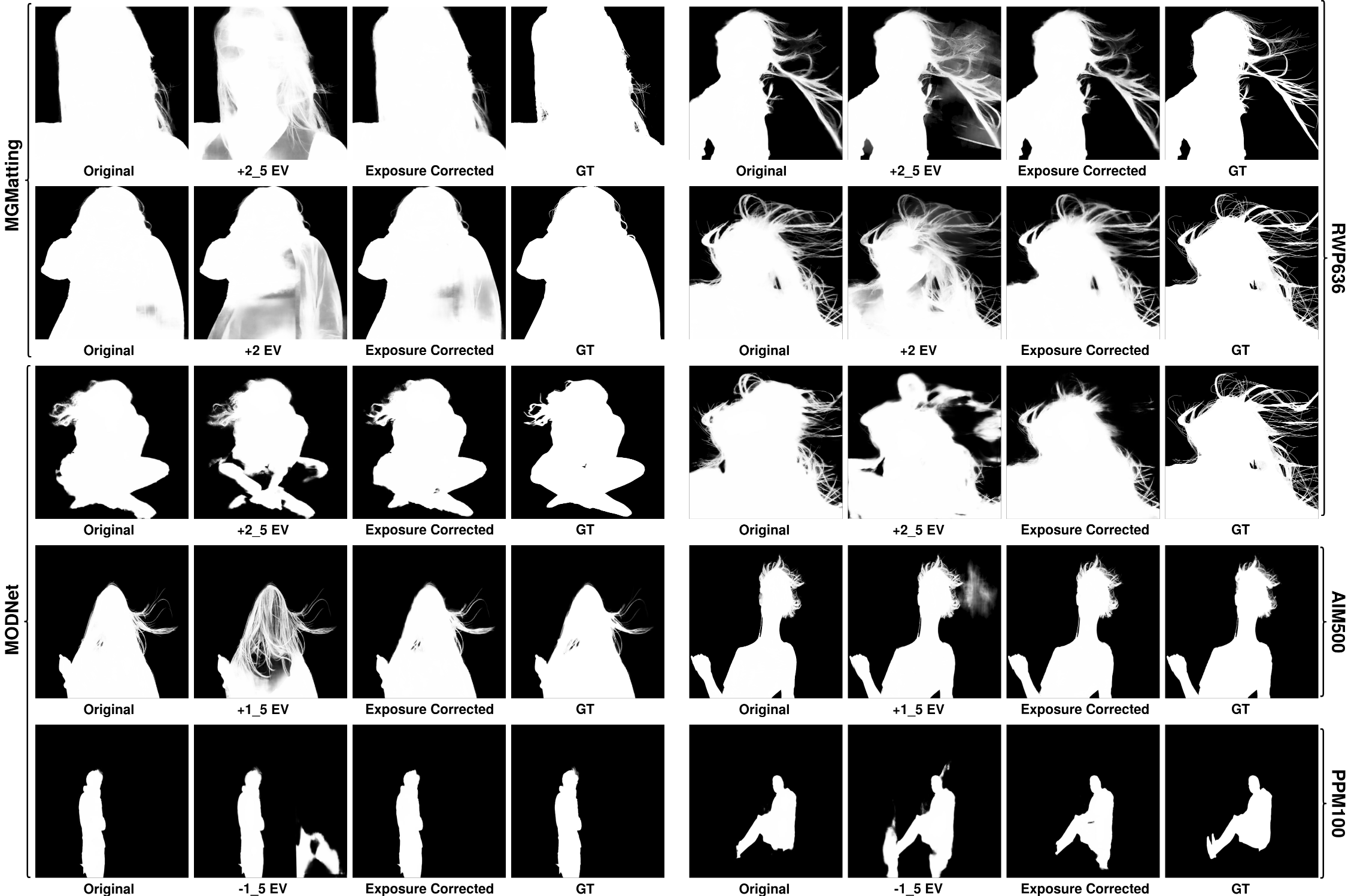}
    \caption{Illustration of the portrait matting outputs. We show sample images from RWP636, AIM500, and PPM100 datasets with MGMatting and MODNet models. The outputs of the remaining models and datasets are the same with this case.} 
    \label{fig:portrait_matting_output}
\end{figure*}

\subsection{Effect of exposure on portrait matting}

Portrait matting models are generally too sensitive to 
the illumination condition of an image. The models have a difficulty to distinguish background scene and foreground subject precisely when the scene is too dark or too bright. Therefore, we perform further experiments in order to investigate the effect of exposure setting, namely well-, under-, and overexposed images, 
on the portrait matting task. 
The experimental results are presented in Table~\ref{tab:portrait_matting}. In the table, letter \textit{E} in the data column indicates the case in which the exposure setting of the images is manipulated with the corresponding EVs, that are presented on the top of the table. Similarly, letter \textit{C} expresses the corrected images by our exposure correction model. In the experiments, we first manipulate the exposure settings of the images to have under- and overexposed images. Then, we perform portrait matting with the well-, under, and overexposed images. Afterward, we employ our exposure correction model, which is trained on the exposure correction dataset~\cite{afifi2021learning}. 
Finally, we test the portrait matting models on these corrected images. 
We also run experiments with the images that no exposure value is applied 
and present the results under \textit{0} column. 


As we can see from the table, the performance of the portrait matting models decreases when the input is under- or overexposed. While the best portrait matting performance is obtained 
with the original images, either by using them directly or by using the corrected version, the models show the worst performance with -2.5 and +2.5 EVs. The performances of the portrait matting models are improved significantly using the corrected versions of the under- and overexposed images. In some cases, the portrait matting models produce a better performance on the corrected version of the original images. This indicates that our exposure correction model enhances even the original images resulting in better portrait matting performance.
Although portrait matting models' performances decrease when the under- and overexposure level increases, the models show almost the same performance after the correction. These results show that our model is remarkably robust against different types and levels of exposure errors. 
The portrait matting models are found to perform better on underexposed images than overexposed images. In other words, overexposed images are more challenging for the portrait matting methods. After the correction, the models show almost the same performance on under- and overexposed images.

The exposure correction dataset~\cite{afifi2021learning} has samples with EVs of -1.5, -1, 0, +1, +1.5. However, in our test set for portrait matting, we also generate images with -2.5, -2, +2, +2.5 EVs. 
In this way, we are able to explicitly explore the performance of our exposure correction model for unseen cases, as 
our model was not trained on them. 
The results show that our model is able to enhance the images outside the training range. 
This demonstrates the generalization capacity of our exposure correction model.

In Figure~\ref{fig:portrait_matting_output}, we show sample results for portrait matting task. In the figure, original column indicates the alpha matte prediction from the original image. The next column demonstrates the predicted results from the input images with different exposure settings. The applied EVs are presented under each image. The third column contains the alpha matte prediction from the corrected version of the under- and overexposed images. Finally, the last column demonstrates the ground truth alpha matte. All these results clearly show the severe degradation in the alpha matte prediction when the input is under- or overexposed image, independent of the employed alpha matte models and datasets. Similarly, when we correct the exposure of the image, the alpha matte prediction performance significantly increases and achieves the similar performance with the original input. This outcome indicates the necessity and usefulness of an exposure correction model as a preprocessing step before the portrait matting. 

\section{Conclusion}

The exposure setting affects the quality and visibility of an image. In this paper, we propose an end-to-end model to address both under- and overexposure problems. We test our model on a large-scale exposure correction dataset and achieve the SOTA results. Besides, we analyze the effect of exposure error on the portrait matting task. For this, we choose three SOTA portrait matting methods and four real-world datasets to perform the experiments. 
Experimental results show that exposure setting affects the portrait matting performance significantly. The proposed exposure correction approach is found to be successful to eliminate the effects of under- and overexposure and able to recover 
the portrait matting performance. In the future work, we will focus on task-specific adaptations to improve the performance as well as increase the stability, especially when the images have under- and overexposed images together. Besides, we will compare different exposure correction models' performance on portrait matting task to compare with our model.






\textbf{Acknowledgement.} The project on which this report is based was funded by the Federal Ministry of Education and Research~(BMBF) of Germany under the number~01IS18040A. 

{\small
\bibliographystyle{ieee_fullname}
\bibliography{paper}
}

\end{document}